\documentclass[conference]{IEEEtran}
\IEEEoverridecommandlockouts
\usepackage{float}
\usepackage[caption=false, font=footnotesize]{subfig}
\usepackage{graphics} 
\usepackage{epsfig} 
\usepackage{mathptmx} 
\usepackage{times} 
\usepackage{amsmath} 
\usepackage{amssymb}  
\usepackage{amsfonts}
\usepackage{cite}
\usepackage{algorithm}
\usepackage{algpseudocode}
\usepackage{multicol}
\usepackage{multirow}
\usepackage{fancyhdr}
\algrenewcommand\algorithmicrequire{\textbf{Input:}}
\algrenewcommand\algorithmicensure{\textbf{Output:}}
\def\BibTeX{{\rm B\kern-.05em{\sc i\kern-.025em b}\kern-.08em
    T\kern-.1667em\lower.7ex\hbox{E}\kern-.125emX}}
\begin{document}

\title{Fast Heuristic Scheduling and Trajectory Planning for Robotic Fruit Harvesters with Multiple Cartesian Arms\\
}

\author{\IEEEauthorblockN{1\textsuperscript{st} Yuankai Zhu}
\IEEEauthorblockA{\textit{Department of Mechanical \& Aerospace Engineering} \\
\textit{University of California, Davis} \\Davis, CA 95616, USA \\
{\tt\small ykzhu@ucdavis.edu}}
\and
\IEEEauthorblockN{2\textsuperscript{nd} Stavros Vougioukas}
\IEEEauthorblockA{\textit{Department of Biological and Agricultural Engineering} \\
\textit{University of California, Davis}\\
Davis, CA 95616, USA \\
{\tt\small svougioukas@ucdavis.edu}}
}

\maketitle
\fancypagestyle{withfooter}{
  \renewcommand{\headrulewidth}{0pt}
  \fancyfoot[C]{\footnotesize Accepted to the Novel Approaches for Precision Agriculture and Forestry with Autonomous Robots IEEE ICRA Workshop - 2025}
}
\thispagestyle{withfooter}
\pagestyle{withfooter}
\begin{abstract}
This work proposes a fast heuristic algorithm for the coupled scheduling and trajectory planning of multiple Cartesian robotic arms harvesting fruits. Our method partitions the workspace, assigns fruit-picking sequences to arms, determines tight and feasible fruit-picking schedules and vehicle travel speed, and generates smooth, collision-free arm trajectories. The fruit-picking throughput achieved by the algorithm was assessed using synthetically generated fruit coordinates and a harvester design featuring up to 12 arms. The throughput increased monotonically as more arms were added. Adding more arms when fruit densities were low resulted in diminishing gains because it took longer to travel from one fruit to another. However, when there were enough fruits, the proposed algorithm achieved a linear speedup as the number of arms increased. 
\end{abstract}


\section{INTRODUCTION}
Fresh fruit harvesting is a labor-intensive operation that relies on a decreasing supply of farm labor\cite{Christiaensen2021work}. Robots are a promising solution to automate harvesting, and researchers have been working on the perception, actuation, and integration aspects of robotic fruit picking for several decades \cite{bac2014harvesting, Kootstra2021review}. The performance metrics that most strongly affect the harvest cost are fruit picking efficiency (FPE), i.e., the ratio of fruits successfully picked over the total number of harvestable fruits on the tree, and fruit picking throughout (FPT), i.e., the average number of fruits picked per unit of time \cite{harrell1987economic}. Given the limitations of how fast a single arm can move and pick without damaging the fruit, the main approach toward increasing FPT has been to use several picking arms. Multi-armed robots with low-cost Cartesian arms that facilitate simpler motion planning and control were proposed early on \cite{edan1993design} and studied in simulation \cite{zion2014harvest, mann2016combinatorial}. In recent years, such harvesters have received renewed attention from academia \cite{arikapudi2021cartesian, li2023multi, arikapudi2023robotic, guo2025dynamic} and industry\footnote{e.g., FFRobotics, Inc.}. Non-Cartesian arms have also been explored by researchers \cite{williams2019robotic, barnett2020work} and startup companies\footnote{e.g., Advanced. Farm, Inc.}. 

\begin{figure}[tb]
    \centering
    \includegraphics[width=0.45\textwidth]{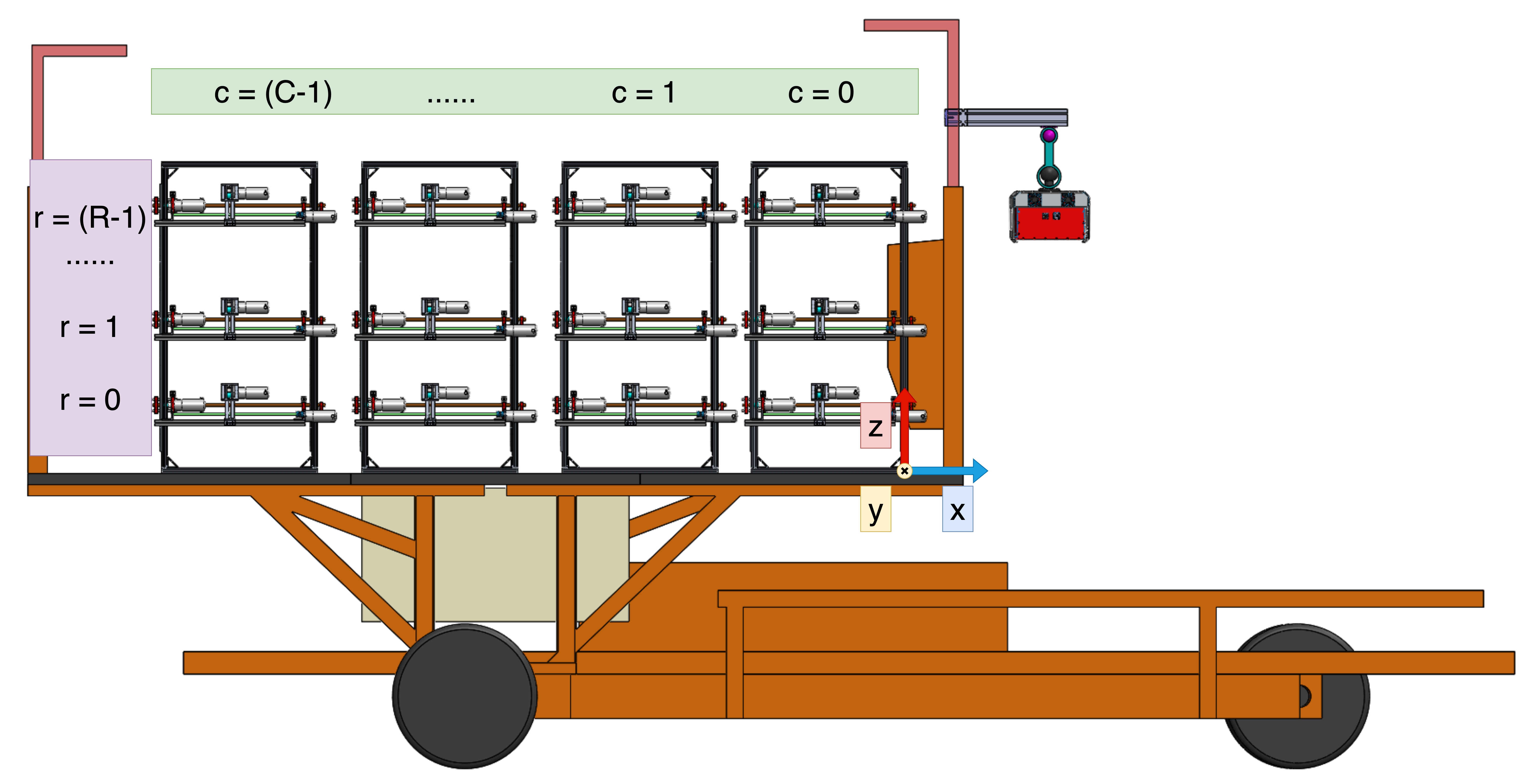}
    \caption{CAD model of a robotic harvester with multiple Cartesian robotic arms on an orchard platform vehicle; groups of arms operate inside rectangular frames.}
    \label{fig: system}
\end{figure}

In this work, we consider a multi-arm Cartesian harvester design with $C$ rectangular frames (also known as cells or columns) and $R$ Cartesian arms inside each cell (Fig. \ref{fig: system}). This harvester is under development and employs a modular architecture with several cells. Inside a cell,  each arm operates independently, constrained only by the cell boundaries and any arms below or above it. The cell size and the number of arms per cell depend on specific harvesting requirements.

\begin{figure}[tb]
    \centering
    \includegraphics[width=0.3\textwidth]{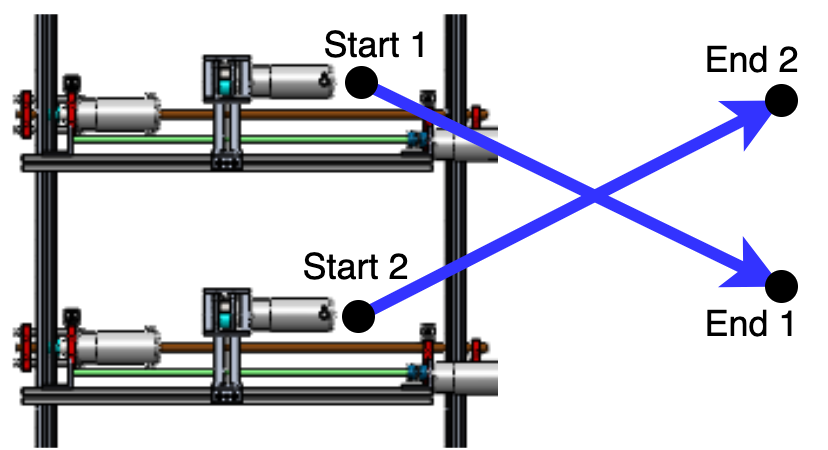}
    \caption{Deadlock scenario: executing a fruit-picking assignment would result in the two arms crossing each other vertically.}
    \label{fig: deadlock}
\end{figure}

Given many fruits and arms, maximizing the FPT requires calculating optimal arm-fruit picking schedules (fruit-picking sequences and timing) that avoid deadlocks (e.g., Fig. \ref{fig: deadlock}) and optimal collision-free trajectories for all arms, along with the vehicle's optimal forward travel speed as the harvester moves along the row. This problem can be considered as the coupling of high-level task scheduling with low-level collision-free arm motion and vehicle trajectory planning. The computational complexities of the task and motion planning problems individually are very hard \cite{garrett2021integrated}. Furthermore, in fruit harvesting, this problem must be solved in real-time. 

In this work, we propose a fast approach to compute deadlock-free fruit-picking schedules, collision-free arm trajectories, and vehicle speeds that lead to high FPTs for the multi-arm harvester under consideration. 


\section{METHODOLOGY} \label{Section: Methodology}


Earlier work \cite{mann2016combinatorial} formulated the multi-arm robotic harvesting problem (MARHP) as a team orienteering problem where the robot maximizes the number of harvested fruits (not necessarily picking them all). Given the high value of fruits, we formulate the MARHP as a minimum-makespan vehicle routing problem (VRP) where all fruits must be picked in the minimum time (maximizing FPT). Our approach assumes 1) all fruit positions are known and can be picked collision-free from branches, 2) arm movement follows trapezoidal motion profiles in all directions, and 3) an arm does not move laterally relative to the fruit while it grasps it. More specifically, when an arm arrives at a fruit in the x-z plane, it extends in the y direction (toward the fruit), detaches the fruit, moves back, and drops it on a conveyor. The extension and retraction motions follow the trapezoidal motion profile with zero initial and end velocity. 

 We assume that a perception system has detected $N$  fruits, and each fruit is assigned a unique index and has known coordinates. For the $r^{th}$ arm in the $c^{th}$ cell, the \textit{arm sequence} $\alpha^{rc}$ is an ordered list of harvested fruit indices, with its corresponding \textit{arm schedule} $S^{rc}$ specifying the arrival times at each fruit in sequence order (e.g., a four-fruit arm sequence is $\alpha^{rc}=\{1,5,7,10\}$, and its schedule is $S^{rc}=\{t_1, t_5, t_7, t_{10}\}$). The sequences/schedules of all arms in a cell form the \textit{cell sequences/schedules} $\alpha_c/S_c$. The cell sequences/schedules of all cells on the machine constitute the \textit{harvester sequences/schedules} $\alpha/S$.

\noindent\textbf{Problem statement} 
\begin{quote}
\textit{Given an initial harvester position, robot arm configurations, and fruit coordinates, determine the vehicle speed $v$, harvester schedule $S$, and collision-free arm trajectories that minimize the makespan $\tau$, subject to arm kinematic constraints.}
\end{quote}

In standard VRP, all travel costs between two nodes can be pre-calculated. However, in MARHP, the travel costs depend on historical sequences, schedules, and vehicle speed. Additionally, two neighboring arms inside a cell can reach a deadlock scenario, as shown in Fig. \ref{fig: deadlock}. 

\begin{figure}[tb]
    \centering
    \includegraphics[width=0.45\textwidth]{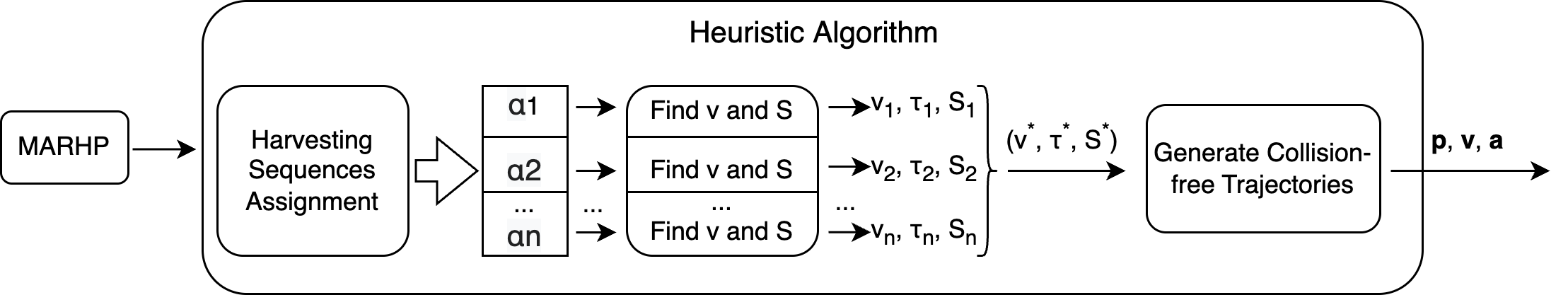}
    \caption{Solution Framework. }
    \label{fig: solution framework}
\end{figure}

Our approach (Fig. \ref{fig: solution framework}) decouples the scheduling and trajectory planning problems by: 1) generating candidate harvesting sequences that prevent deadlock, 2) computing the best vehicle moving speed and the corresponding schedule for each sequence by iteratively updating the arrival times at each fruit - considering the collision risk in both horizontal and vertical directions, - choosing the solution with the minimum makespan, and 3) generating smooth collision-free trajectories for the best candidate harvesting sequences under their vehicle speed and schedules. 

\subsection{Generate Fruit-Picking Arm Sequences} \label{subsection: assign the harvesting sequences}

\begin{figure}[tb]
    \centering
    \subfloat[Step 1: create intervals.]{
        \centering
        \includegraphics[width=0.21\textwidth]{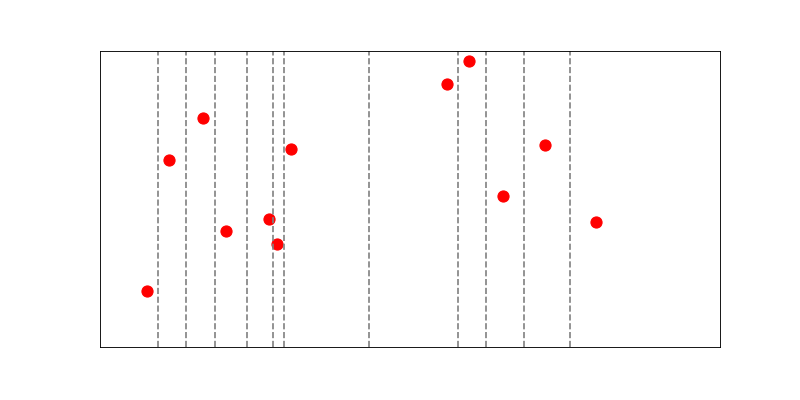}
        \label{fig: step 1}
    }\quad
    \subfloat[Step 2: merge intervals.]{
        \centering
        \includegraphics[width=0.21\textwidth]{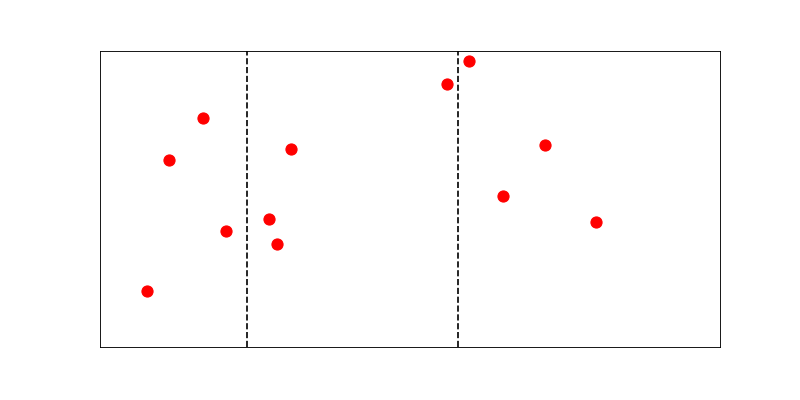}
        \label{fig: step 2}
    }\\
    \subfloat[Step 3: split into clusters.]{
        \centering
        \includegraphics[width=0.21\textwidth]{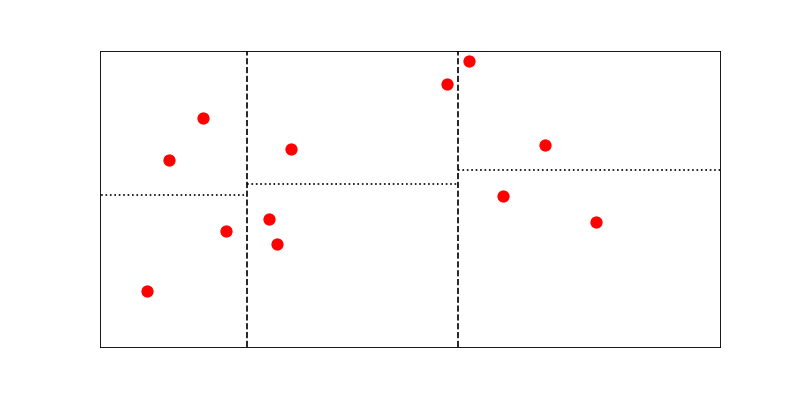}
        \label{fig: step 3}
    }\quad
    \subfloat[Step 4: choose one fruit from each cluster to form the harvester sequences. The cell sequences are in the same color.]{
        \centering
        \includegraphics[width=0.21\textwidth]{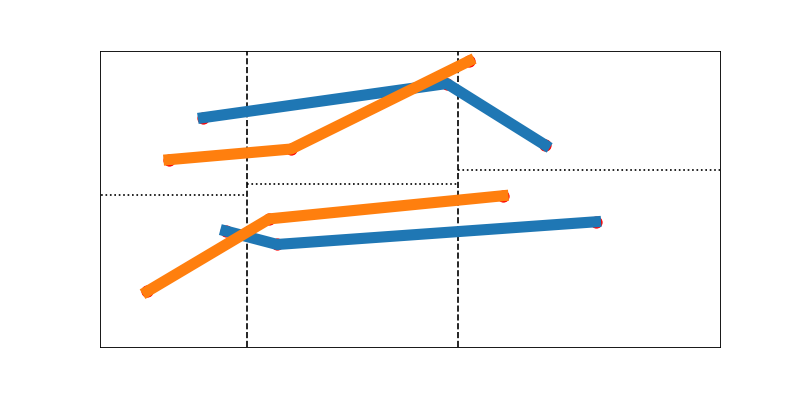}
        \label{fig: step 4}
    }
    \caption{Example of fruit clustering and sequence generation for a robot with 2 cells and 2 arms per cell.}
    \label{fig: fruit allocation}
\end{figure}

In the VRP, balancing the workload among agents is often done by partitioning the graph and assigning independent sub-graphs to individual agents. We use the same approach and propose a fruit clustering and assignment method to balance the workload distribution of the arms. The method comprises the following steps (Fig. \ref{fig: fruit allocation}):
\begin{enumerate}
    \item Create vertical zones - one per fruit - by computing the midpoint line between successive fruits.\label{heuristic step 1}
    \item Merge vertical zones into wider zones that contain $m=R\cdot C$ fruits (the last zone may have fewer fruits). \label{heuristic step 2}
    \item Split each zone into $R$ clusters/cells, each containing $C$ fruits. \label{heuristic step 3}
    \item For each arm, generate two candidate picking sequences by selecting the topmost or rightmost available fruit in each cluster. The total number of candidate sequences is $n=2m$.\label{heuristic step 4}
\end{enumerate}

\subsection{Calculate Travel Speed and Sequence Timing} \label{subsection: determine the scheduling}
If the vehicle's travel speed is too slow, the arms may finish harvesting all available fruits and remain idle until more fruits enter their workspace. In contrast, if the vehicle speed is too fast, an arm may not have enough time to pick some fruits. We use binary search to find the maximum feasible travel speed $v$ and picking schedules that result in the minimum makespan. Algorithm \ref{alg: binary search algorithm} performs a binary search for the speed (outer loop) while calculating the corresponding fruit-picking timing information for each picking sequence candidate (inner loop).

\begin{algorithm}[tb]
\caption{Find Velocity v and Schedules S}
\begin{algorithmic}[1]
\Require $\alpha$
\State $v_{low}=0, v_{high}=\frac{l}{\max\{\text{Tp}_i\}}$, $v, \tau, S$ \label{alg line: init}
\While{$v_{high} - v_{low} > \epsilon$} \label{alg line: start while}
    \State $v_{mid} = (v_{high} + v_{low})/2$ \label{alg line: find middle}
    \State $f=true, \tau'=0, S'=\{\}$ \label{alg line: init sub}
    \For {$\alpha_c$ of $\alpha$} \label{alg line: start for loop}
    \State $f_c, \tau_c, S_c = \mathbf{cellSchedulingAtV}(\alpha_c, v_{mid})$\label{alg line: find cell info}
    \If{$f_c$} \label{alg line: check cell feasibility}
        \State $\tau'=\max(\tau', \tau_c)$, $S'.push(S_c)$ \label{alg line: update cell info}
    \Else \label{alg line: check cell feasibility else}
        \State $f = false$ \label{alg line: update feasibility}
        \State break \label{alg line: break}
    \EndIf \label{alg line: check cell feasibility end}
    \EndFor \label{alg line: end for loop}
    \If {$f$} \label{alg line: check feasibility}
        \State $v_{low} = v_{mid}$ \label{alg line: check larger velocity}
        \State $v = v_{mid}$, $\tau = \tau'$, $S = S'$ \label{alg line: update current best info}
    \Else \label{alg line: check feasibility else}
        \State $v_{high} = v_{mid}$ \label{alg line: check smaller velocity}
    \EndIf \label{alg line: check feasibility end}
\EndWhile \label{alg line: end while}
\Ensure $v$, $\tau$, $S$
\end{algorithmic}
\label{alg: binary search algorithm}
\end{algorithm}


Starting at time $0$ and an initial position, the arrival times of each arm at the  fruits in a candidate sequence must be calculated, while ensuring that the arm does not hit the cell frame or another arm (above or below). This calculation is performed by Algorithm \ref{alg: cell scheduling under v}.

\begin{algorithm}[tb]
\caption{Find Cell Schedule (\textbf{cellSchedulingAtV})}
\begin{algorithmic}[1]
\Require $\alpha_c$, $v$
\State $t_{cur}[R]=\mathbf{0}$, $S_c=\mathbf{0}$, $\textit{YieldInfo}[R]=\{\}$\label{alg line: init cur time, schedule and yielding info}
\For {$k$, $x_{cur}[R]$, $z_{cur}[R]$, $x_{next}[R]$, $z_{next}[R]$ of $\alpha_c$}\label{alg line: iteration start}
    \For {$r = 1:R$} \label{alg line: update yield info start}
        \State $\mathbf{Update}$(\textit{YieldInfo[$r$]}, $r$, $z_{cur}[R]$, $t_{cur}[R]$);
    \EndFor \label{alg line: update yield info end}
    \For {$r = 1:R$} \label{alg line: update schedule start}
        \State \textit{vTime} $= \mathbf{VT}(\textit{YieldInfo}[r], z_{cur}[r], z_{next}[r], t_{cur}[r])$\label{alg line: vertical time}
        \State \textit{hTime} $= \mathbf{HT}(x_{cur}[r], x_{next}[r], t_{cur}[r], v)$\label{alg line: horizontal time}
        \State $t_{next}=\max\{\textit{vTime}, \textit{hTime}\}$\label{alg line: update arriving time}
        \State $S_c[r][k]=t_{next}$\label{alg line: update schedules}
        \State $t_{cur}[r] = t_{next}+\text{Tp}_{\alpha_c[r][k]}$\label{alg line: update start moving time}
        \If{Arm[r] hits the rear end of the cell} \label{alg line: check hitting}
            \State \textbf{Output:} false, null, null
        \EndIf \label{alg line: check hitting end}
    \EndFor \label{alg line: update schedule end}
\EndFor \label{alg line: iteration end}
\State $\tau_c = \max\{t_{cur}[R]\}$\label{alg line: update cell makespan}
\Ensure true, $\tau_c$, $S_c$
\end{algorithmic}
\label{alg: cell scheduling under v}
\end{algorithm}

The inputs are the current cell harvesting sequences $\alpha_c$ (each contains $R$ arm sequences), and the vehicle travel speed $v$. The algorithm caclulates the time required for each arm to reach the next fruit using a trapezoidal speed profile (Algorithm \ref{alg: vertical time}, Algorithm  \ref{alg: horizontal time}), and incorporating any time the arm must wait to $yield$ for another arm to move and avoid collisions (Algorithm \ref{alg: update yielding information}).

\begin{algorithm}[tb]
\caption{Update yielding Information (\textbf{Update})}
\begin{algorithmic}[1]
\Require \textit{CurYieldInfo}, $r$, $z_{cur}[R]$, $t_{cur}[R]$;
\State $r_{yield} = r$ \label{alg line: start from current arm}
\State $\textit{direction} =$ arm $r$ vertical moving direction ($\pm1$)\label{alg line: find moving direction}
\For {$r_{yield} = r_{yield} + \textit{direction}$ and $1 \leq r_{yield} \leq R$}\label{alg line: check along the moving direction}
\If {Current arm $r$ needs to yield arm $r_{yield}$} \label{alg line: yielding condition}
\State \textit{CurYieldInfo}.add($z_{cur}$[$r_{yield}$], $t_{cur}$[$r_{yield}$])\label{alg line: update yielding information}
\EndIf
\EndFor
\end{algorithmic}
\label{alg: update yielding information}
\end{algorithm}




\begin{algorithm}[tb]
\caption{Vertical Time (\textbf{VT})}
\begin{algorithmic}[1]
\Require \textit{YieldInfo}, $z_{cur}$, $z_{target}$ $t$
\For{$z_{yield}$, $t_{yield}$ of \textit{YieldInfo}}\label{alg line: iterate yielding info}
\State $\Delta z = |z_{yield} - z_{cur}|$
\State $\Delta t = \mathbf{Tr}(\Delta z, v_{max}, a_{max})$
\State $t = \max\{t_{yield}, t+\Delta t\}$\label{alg line: update time at yielding points}
\EndFor
\State $\Delta z = |z_{target} - z_{yield}|$
\State $\Delta t = \mathbf{Tr}(\Delta z, v_{max}, a_{max})$
\State $t = t + \Delta t$\label{alg line: update time at target}
\Ensure $t$
\end{algorithmic}
\label{alg: vertical time}
\end{algorithm}

\begin{algorithm}[tb]
\caption{Horizontal Time (\textbf{HT})}
\begin{algorithmic}[1]
\Require $x_{cur}$, $x_{target}$, $t$, $v$
\State $T = \mathbf{Tr}(x_{target}-x_{cur}, v_{max}+v, a_{max})$;
\State $d_f = x_{frontInit} + v\cdot t - x_{cur}$\label{alg line: distance to front end}
\If{$d_f \geq S2 - S1$}\label{alg line: check condition hitting front end}
\State $t = t + T$\label{alg line: full speed and accelerate HT}
\Else
\State $t = \frac{x_{target}-x_{frontInit}}{v} + \frac{v}{2a_{max}}$\label{alg line: front end limit HT}
\EndIf
\Ensure $t$
\end{algorithmic}
\label{alg: horizontal time}
\end{algorithm}

\subsection{Generate Collision-free Trajectories} \label{subsection: generate trajectory}

Given the travel speed $v$ and the harvester schedules $S^{rc}_k$, trajectory generation is cast as a MILP problem.  The problem constants are defined in Table \ref{table: constants of trajs}, and the variables in Table \ref{table: variables of trajs}. The indices used are the row index $r\in \{1,...,R\}$, cell index $c\in \{1,...,C\}$, time stamp $ts\in \{0,1,2,...,\tau\}$, fruit index $i\in \{1,...,N\}$, and sequence order $\alpha^{rc}_k\in \{1,...,N\}$. 

\begin{table}[tb]
\caption{Trajectories Generation Constants Definitions}
\label{table: constants of trajs}
\begin{center}
\resizebox{\linewidth}{!}{%
\begin{tabular}{|l|l|l|}
\hline
\textbf{Constants} & \textbf{Definition}                            & \textbf{Note} \\ \hline
$\tau$            & Makespan                        &  Indexed with $ts$     \\ \hline
$dt$                & Time step interval &               \\ \hline
$v$      & Vehicle speed & \\ \hline
l                 & Length of each cell                          &               \\ \hline
H                 & Height of frame &               \\ \hline
R                 & Number of arms in one cell                   &   Indexed with $r$     \\ \hline
C                 & Number of cells of the robot             & Indexed with $c$     \\ \hline
N                 & Number of fruits in total                      &  Indexed with $i$  \\ \hline
$\text{Tp}_i$                & Picking duration for fruit i                &               \\ \hline
$\alpha^{rc}_k$ & Arm ($r$, $c$) sequence & Indexed with $k$ \\ \hline
$S_k^{rc}$ & Schedule of arm (r, c) sequence & \\ \hline
$\mathbf{f}_i = [fx_i, fz_i]$                 &Position of fruit i             &   \\ \hline
$v_{max}$           & Max speed of the arm in one direction&               \\ \hline
$a_{max}$            & Max acceleration of the arm in one direction&               \\ \hline
$\mathbf{p}^{rc}_0$, $\mathbf{v}^{rc}_0$, $\mathbf{a}^{rc}_0$  & Initial conditions & \\ \hline
\end{tabular}
}
\end{center}
\end{table}

\begin{table}[tb]
\caption{Trajectories Generation Variables Definitions}
\centering
\resizebox{\linewidth}{!}{%
\begin{tabular}{|l|l|l|}
\hline
\textbf{Variables} & \textbf{Definition} & \textbf{Note} \\ \hline
$\mathbf{p}^{rc}[ts] = [p^{rc}_x[ts], p^{rc}_z[ts]]$ &Position of arm (r, c) at ts & Continuous \\ \hline
$\mathbf{v}^{rc}[ts] = [v^{rc}_x[ts], v^{rc}_z[ts]]$ & Velocity of arm (r, c) at ts & Continuous \\ \hline
$\mathbf{a}^{rc}[ts] = [a^{rc}_x[ts], a^{rc}_z[ts]]$ & Acceleration of arm (r, c) at ts & Continuous \\ \hline
\end{tabular}%
}
\label{table: variables of trajs}
\end{table}

\noindent
\textbf{Trajectory Generation}
\begin{equation}
\min \sum^{R}_{r=1}\sum^{C}_{c=1}\sum^{\tau}_{ts=1}(|a^{rc}_x[ts]|+|a^{rc}_z[ts]|) \label{objective: minimum absolute acceleration sums} \nonumber \\
\end{equation}
\text{Subject to:}\\
\begin{subequations}
\begin{align}
& \mathbf{p}^{rc}[0] = \mathbf{p}^{rc}_0, \mathbf{v}^{rc}[0]=\mathbf{v}^{rc}_0 \label{constraints: initial conditions} \\
& \mathbf{v}^{rc}[ts+1] = \mathbf{v}^{rc}[ts] + \mathbf{a}^{rc}[ts]\cdot dt
\label{constraints: velocity waypoints update} \\
& \mathbf{p}^{rc}[ts+1] = \mathbf{p}^{rc}[ts] + \mathbf{v}^{rc}[ts]\cdot dt + 0.5\mathbf{a}^{rc}[ts]\cdot dt^2
\label{constraints: acceleration waypoints update} \\
& -v_{max}+v \leq v^{rc}_x[ts] \leq v_{max}+v \label{constraints: x velocity constraints} \\
& -v_{max} \leq v^{rc}_z[ts] \leq v_{max} \label{constraints: z velocity constraints} \\
& -a_{max} \leq a^{rc}_x[ts], a^{rc}_z[ts] \leq a_{max} \label{constraints: acceleration constraints} \\
& p^{rc}_z[ts] \leq p^{r+1,c}_z[ts]\label{constraints: collision avoidance} \\
& -c \cdot l + v\cdot ts \cdot dt \leq p^{rc}_x[ts] \leq -(c-1) \cdot l + v\cdot ts \cdot dt \label{constraints: waypoint horizontal bounds} \\
& 0 \leq p^{0c}_z[ts] \label{constraints: waypoint vertical lower bounds} \\
& p^{Rc}_z[ts] \leq H \label{constraints: waypoint vertical upper bounds} \\
& \mathbf{p}^{rc}[ts] = \mathbf{f}_{\alpha^{rc}_k},\ \text{when}\ S^{rc}_k \leq ts\cdot dt \leq S^{rc}_k + \text{Tp}_{\alpha^{rc}_k} \label{constraints: staying constraints}
\end{align}
\end{subequations}

This MILP model minimizes the total absolute accelerations under the constraints (a) to (k): (\ref{constraints: initial conditions}) defines the initial conditions; (\ref{constraints: velocity waypoints update}) and (\ref{constraints: acceleration waypoints update}) are the waypoints update constraints; (\ref{constraints: x velocity constraints}), (\ref{constraints: z velocity constraints}) and (\ref{constraints: acceleration constraints}) are the kinematic constraints; (\ref{constraints: collision avoidance}) guarantees the arms in the same cell don't collide each other; (\ref{constraints: waypoint horizontal bounds}) defines the horizontal bounds of waypoints; (\ref{constraints: waypoint vertical lower bounds}) and (\ref{constraints: waypoint vertical upper bounds}) define the vertical bounds of waypoints; (\ref{constraints: staying constraints}) ensures the arm stays at the fruits during the picking process.

\section{EXPERIMENTAL RESULTS} \label{Section: Simulation results}
\begin{figure}[tb]
    \centering
    \subfloat[Uniform.]{
        \includegraphics[width=0.4\textwidth]{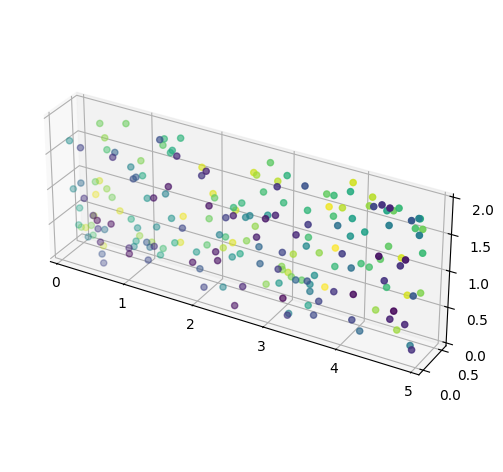}
        \label{fig: uniform distribution}
    }
    
    \caption{Example of an orchard row segment with a randomly generated uniform fruit distribution (color varies with depth).}
    \label{fig: fruits distributions}
\end{figure}

We evaluated the FPT ($N/\tau$) achieved by the proposed algorithms with synthetic data sets that contained uniformly distributed fruits over a length of $50 m$, a height of $2.0 m$ and canopy depth of $0.5m$ (Fig. \ref{fig: uniform distribution}). The $(R, C)$ combinations were $(1,1)$, $(2,1)$, $(3,1)$, $(3,2)$, $(3,3)$, $(3,4)$, and corresponded to $1$, $2$, $3$, $6$, $9$, and $12$ arms. The fruit distribution densities were  $\rho=5,10,30,100\;\text{fruits}/m^2$. For each case, we ran 100 harvest tests with randomly generated fruit coordinates. We used the following parameters: $dt = 0.1s$, $T_d=2s$, $l = 0.6m$, $H = 2m$, $v_{max}=1m/s$, $a_{max}=1m/s^2$. All tests ran on a desktop with an AMD Ryzen 9 7950X 16-core processor and 128Gb memory. The MILP was solved using the Gurobi optimizer\cite{gurobi}. Example harvest animations can be found at the following link (videos\footnote{https://bitbucket.org/bioautomationlab/simulation-videos/}).

\begin{figure}[t]
    \centering
    \includegraphics[width=0.4\textwidth]{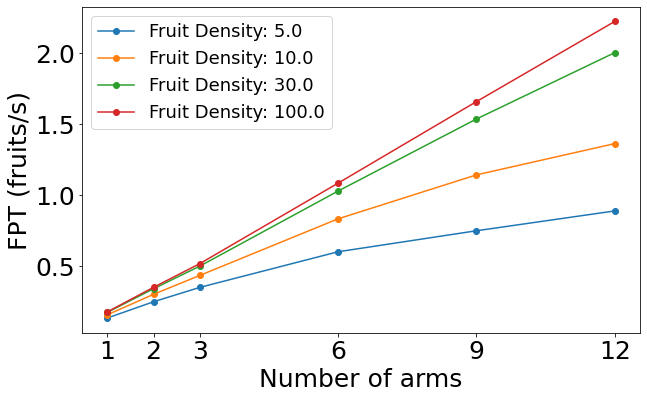}
    \caption{Each point represents the average FPT over 100 randomly-generated fruit distributions.}
    \label{fig: number of arms vs results for different density of fruits}
\end{figure}

 The FPTs are shown in Fig \ref{fig: number of arms vs results for different density of fruits}. The FPT increased monotonically as more arms were added. Adding more arms when fruit densities were low resulted in diminishing gains because it took longer to travel from one fruit to another. However, when there were enough fruits, the proposed algorithm achieved a linear speedup as the number of arms increased. When the fruit density was 100 $\text{fruits}/m^2$, the harvester picked 10,000 fruits. When one arm was used, the FPT was 0.17 fruits/s, whereas with 12 arms, the FPT was 2.21 fruits/s, approximately 12.7 times faster.

\section{CONCLUSION AND FUTURE WORK} \label{Section: Future work}
A heuristic algorithm was developed to solve MARHP in seconds for long orchard segments with thousands of fruits. The algorithm can achieve linear speedup in fruit picking throughput as the number of arms increases, thus promising to lead to machines that can harvest many times faster than a human. The algorithm will be extended to perform dynamic planning, since orchard rows are long and fruit distributions can be reliably estimated (using cameras) over limited length horizons in front of a harvester. A prototype of this multi-arm harvester design is under development and will be used to test the approach in real-world conditions.

\bibliographystyle{ieeetr}
\bibliography{Bibliography}

\end{document}